\documentclass[11pt,a4paper]{article}
\usepackage{acl2015}
\usepackage{times}
\usepackage{url}
\usepackage{latexsym}
\usepackage{amsmath}
\usepackage{graphicx}
\usepackage{color}
\usepackage{rotating}
\usepackage{multirow}

\usepackage{array}
\newcolumntype{L}[1]{>{\raggedright\let\newline\\\arraybackslash\hspace{0pt}}m{#1}}
\newcolumntype{C}[1]{>{\centering\let\newline\\\arraybackslash\hspace{0pt}}m{#1}}
\newcolumntype{R}[1]{>{\raggedleft\let\newline\\\arraybackslash\hspace{0pt}}m{#1}}

\title{Syntax-Aware Multi-Sense Word Embeddings\\for Deep Compositional Models of Meaning} 

\author{Jianpeng Cheng \\
 University of Oxford \\
 Department of \\ Computer Science \\
 {\tt\fontsize{11pt}{11pt}\selectfont jianpeng.cheng@stcatz.oxon.org} \\
 \And
 Dimitri Kartsaklis \\
 Queen Mary University of London \\
 School of Electronic Engineering \\
 and Computer Science \\
 {\tt\fontsize{11pt}{11pt}\selectfont d.kartsaklis@qmul.ac.uk} \\
 }

\def\bR{\begin{color}{red}} 
\def\bB{\begin{color}{blue}}
\def\e{\end{color}}

\begin{document}

\maketitle

\begin{abstract}
Deep compositional models of meaning acting on distributional representations of words in order to produce vectors of larger text constituents are evolving to a popular area of NLP research. We detail a compositional distributional framework based on a rich form of word embeddings that aims at facilitating the interactions between words in the context of a sentence. Embeddings and composition layers are jointly learned against a generic objective that enhances the vectors with syntactic information from the surrounding context. Furthermore, each word is associated with a number of senses, the most plausible of which is selected dynamically during the composition process. We evaluate the produced vectors qualitatively and quantitatively with positive results. At the sentence level, the effectiveness of the framework is demonstrated on the MSRPar task, for which we report results within the state-of-the-art range.
\end{abstract}

\section{Introduction}

Representing the meaning of words by using their distributional behaviour in a large text corpus is a well-established technique in NLP research that has been proved useful in numerous tasks. In a distributional model of meaning, the semantic representation of a word is given as a vector in some high dimensional vector space, obtained either by explicitly collecting co-occurrence statistics of the target word with words belonging to a representative subset of the vocabulary, or by directly optimizing the word vectors against an objective function in some neural-network based architecture \cite{collobert2008unified,mikolov2013distributed}. 

Regardless their method of construction, distributional models of meaning do not scale up to larger text constituents such as phrases or sentences, since the uniqueness of multi-word expressions would inevitably lead to data sparsity problems, thus to unreliable vectorial representations. The problem is usually addressed by the provision of a compositional function, the purpose of which is to prepare a vectorial representation for a phrase or sentence by combining the vectors of the words therein. While the nature and complexity of these compositional models may vary, approaches based on deep-learning architectures have been shown to be especially successful in modelling the meaning of sentences for a variety of tasks \cite{socher2012,KalchbrennerACL2014}.

The mutual interaction of distributional word vectors by a means of a compositional model provides many opportunities for interesting research,  the majority of which still remains to be explored. One such direction is to investigate in what way lexical ambiguity affects the compositional process. In fact, recent work has shown that shallow multi-linear compositional models that explicitly handle extreme cases of lexical ambiguity in a step prior to composition present consistently better performance than their ``ambiguous'' counterparts \cite{kartsaklis:2013:EMNLP,KartsaklisACL}. A first attempt to test these observations in a deep compositional setting has been presented by \newcite{cheng2014} with promising results.

Furthermore, a second important question relates to the very nature of the word embeddings used in the context of a compositional model. In a setting of this form, word vectors are not any more just a means for discriminating words based on their underlying semantic relationships; the main goal of a word vector is to contribute to a bigger whole---a task in which syntax, along with semantics, also plays a very important role. It is a central point of this paper, therefore, that in a compositional distributional model of meaning word vectors should be injected with information that reflects their syntactical roles in the training corpus.

The purpose of this work is to improve the current practice in deep compositional models of meaning in relation to both the compositional process itself and the quality of the word embeddings used therein. We propose an architecture for jointly training a compositional model and a set of word embeddings, in a way that imposes dynamic word sense induction for each word during the learning process. 
Note that this is in contrast with recent work in multi-sense neural word embeddings \cite{neelakantan2014efficient}, in which the word senses are learned without any compositional considerations in mind. 

Furthermore, we make the word embeddings syntax-aware by introducing a variation of the hinge loss objective function of \newcite{collobert2008unified}, in which the goal is not only to predict the occurrence of a target word in a context, but to also predict the \textit{position} of the word within that context. A qualitative analysis shows that our vectors reflect both semantic and syntactic features in a concise way. 

In all current deep compositional distributional settings, the word embeddings are internal parameters of the model with no use for any other purpose than the task for which they were specifically trained. In this work, one of our main considerations is that the joint training step should be generic enough to not be tied in any particular task. In this way the word embeddings and the derived compositional model can be learned on data much more diverse than any task-specific dataset, reflecting a wider range of linguistic features. Indeed, experimental evaluation shows that the produced word embeddings can serve as a high quality general-purpose semantic word space, presenting performance on the Stanford Contextual Word Similarity (SCWS) dataset of \newcite{huang2012improving} competitive to and even better of the performance of well-established neural word embeddings sets.

Finally, we propose a dynamic disambiguation framework for a number of existing deep compositional models of meaning, in which the multi-sense word embeddings and the compositional model of the original training step are further refined according to the purposes of a specific task at hand. In the context of paraphrase detection, we achieve a result very close to the current state-of-the-art on the Microsoft Research Paraphrase Corpus \cite{dolan2005automatically}. 
An interesting aspect at the sideline of the paraphrase detection experiment is that, in contrast to mainstream approaches that mainly rely on simple forms of classifiers, we approach the problem by following a siamese architecture \cite{bromley1993signature}.

\section{Background and related work}


\subsection{Distributional models of meaning}

Distributional models of meaning follow the \textit{distributional hypothesis} \cite{harris1954distributional}, which states that two words that occur in similar contexts have similar meanings. Traditional approaches for constructing a word space rely on simple counting: a word is represented by a vector of numbers (usually smoothed by the application of some function such as point-wise mutual information) which show how frequently this word co-occurs with other possible context words in a corpus of text.

In contrast to these methods, a recent class of distributional models treat word representations as parameters directly optimized on a word prediction task \cite{bengio2003neural,collobert2008unified,mikolov2013distributed,pennington2014glove}.   Instead of relying on observed co-occurrence counts, these models aim to maximize the objective function of a neural net-based architecture; \newcite{mikolov2013distributed}, for example, compute the conditional probability of observing words in a context around a target word (an approach known as the {\em skip-gram model}). Recent studies have shown that, compared to their co-occurrence counterparts, neural word vectors reflect better the semantic relationships between words \cite{baroni2014don} and are more effective in compositional settings \cite{milajevs2014}. 

\subsection{Syntactic awareness}

Since the main purpose of distributional models until now was to measure the semantic relatedness of words, relatively little effort has been put into making word vectors aware of information regarding the syntactic role under which a word occurs in a sentence. In some cases the vectors are POS-tag specific, so that `book' as noun and `book' as verb are represented by different vectors \cite{kartsaklis:2013:EMNLP}. Furthermore, word spaces in which the context of a target word is determined by means of grammatical dependencies \cite{pado2007} are more effective in capturing syntactic relations than approaches based on simple word proximity. 

For word embeddings trained in neural settings, syntactic information is not usually taken explicitly into account, with some notable exceptions. At the lexical level, \newcite{levy-2014} propose an extension of the skip-gram model based on grammatical dependencies. Following a different approach, \newcite{mnih-2013} weight the vector of each context word depending on its distance from the target word. With regard to compositional settings (discussed in the next section), \newcite{hashimoto-2014} use dependency-based word embeddings by employing a hinge loss objective, while \newcite{hermann2013} condition their objectives on the CCG types of the involved words. 

As we will see in Section \ref{sec:syntax}, the current paper offers an appealing alternative to those approaches that does not depend on grammatical relations or types of any form.

\subsection{Compositionality in distributional models}
\label{sec:compdis}

The methods that aim to equip distributional models of meaning with compositional abilities come in many different levels of sophistication, from simple element-wise vector operators such as addition and multiplication \cite{mitchell-lapata:2008:ACLMain} to category theory \cite{coecke2010mathematical}. In this latter work relational words (such as verbs or adjectives) are represented as multi-linear maps acting on vectors representing their arguments (nouns and noun phrases). In general, the above models are shallow in the sense that they do not have functional parameters and the output is produced by the direct interaction of the inputs; yet they have been shown to capture the compositional meaning of sentences to an adequate degree.

The idea of using neural networks for compositionality in language appeared 25 years ago in a seminal paper by \newcite{pollack1990}, and has been recently re-popularized by Socher and colleagues \cite{socher2011dynamic,socher2012}. The compositional architecture used in these works is that of a \textit{recursive neural network} (RecNN) \cite{socher2011parsing}, where the words get composed by following a parse tree. A particular variant of the RecNN is the \textit{recurrent neural network} (RNN), in which a sentence is assumed to be generated by aggregating words in sequence \cite{mikolov2010recurrent}. Furthermore, some recent work \cite{KalchbrennerACL2014} models the meaning of sentences by utilizing the concept of a convolutional neural network \cite{lecun1998gradient}, the main characteristic of which is that it acts on small overlapping parts of the input vectors. In all the above models, the word embeddings and the weights of the compositional layers are optimized against a task-specific objective function. 

In Section \ref{sec:syntax} we will show how to remove the restriction of a supervised setting, introducing a generic objective that can be trained on any general-purpose text corpus. While we focus on recursive and recurrent neural network architectures, the general ideas we will discuss are in principle model-independent.

\subsection{Disambiguation in composition}  

Regardless of the way they address composition, all the models of Section \ref{sec:compdis} rely on ambiguous word spaces, in which every meaning of a polysemous word is merged into a single vector. Especially for cases of homonymy (such as `bank', `organ' and so on), where the same word is used to describe two or more completely unrelated concepts, this approach is problematic: the semantic representation of the word becomes the average of all senses, inadequate to express any of them in a reliable way. 

To address this problem, a prior disambiguation step on the word vectors is often introduced, the purpose of which is to find the word representations that best fit to the given context, before composition takes place \cite{reddy2011dynamic,karts_conll2013,kartsaklis:2013:EMNLP,KartsaklisACL}. This idea has been tested on algebraic and tensor-based compositional functions with very positive results. Furthermore, it has been also found to provide minimal benefits for a RecNN compositional architecture in a number of phrase and sentence similarity tasks \cite{cheng2014}. This latter work clearly suggests that explicitly dealing with lexical ambiguity in a deep compositional setting is an idea that is worth to be further explored. While treating disambiguation as only a preprocessing step is a strategy less than optimal for a neural setting, one would expect that the benefits should be  greater for an architecture in which the disambiguation takes place in a dynamic fashion during training. 

We are now ready to start detailing a compositional model that takes into account the above considerations. The issue of lexical ambiguity is covered in Section \ref{sec:senses}; Section \ref{sec:syntax} below deals with generic training and syntactic awareness. 

\section{Syntax-based generic training}
\label{sec:syntax}

We propose a novel architecture for learning word embeddings and a compositional model to use them in a single step. The learning takes places in the context of a RecNN (or an RNN), and both word embeddings and parameters of the compositional layer are optimized against a generic objective function that uses a hinge loss function. 

\begin{figure}[t!]
   \centering
   \includegraphics[scale=0.31]{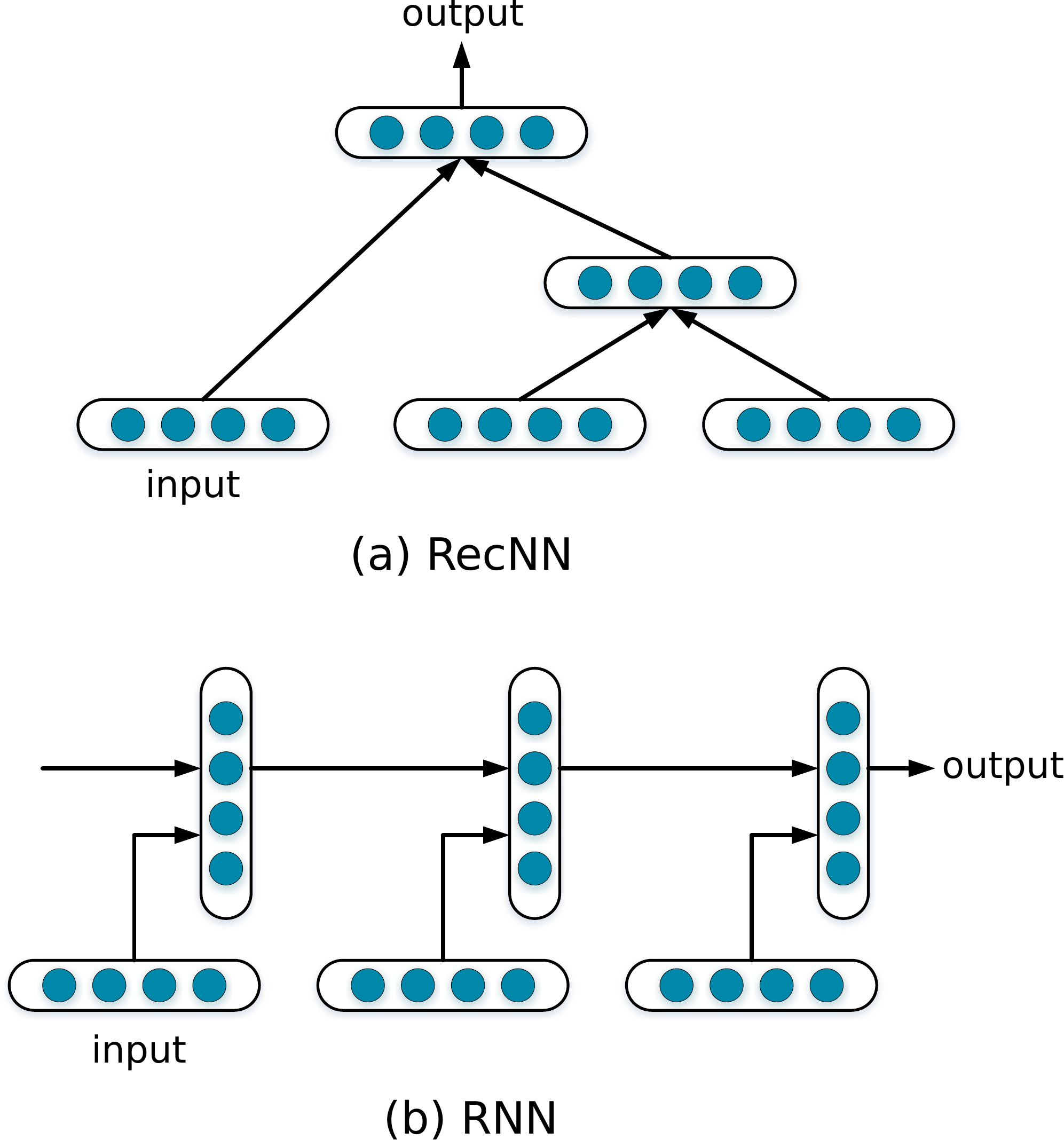}
\caption{Recursive (a) and recurrent (b) neural networks.}
\label{fig:recnn}	
\end{figure}

Figure \ref{fig:recnn} shows the general form of recursive and recurrent neural networks. In architectures of this form, a compositional layer is applied on each pair of inputs $\mathbf{x_1}$ and $\mathbf{x_2}$ in the following way:

\vspace{-0.3cm}
\begin{equation}
\mathbf{p} = g(\mathbf{W}\mathbf{x_{[1:2]}} + \mathbf{b})
\end{equation}

\noindent where $\mathbf{x_{[1:2]}}$ denotes the concatenation of the two vectors, $g$ is a non-linear function, and $\mathbf{W},\mathbf{b}$ are the parameters of the model. In the RecNN case, the compositional process continues recursively by following a parse tree until a vector for the whole sentence or phrase is produced; on the other hand, an RNN assumes that a sentence is generated in a left-to-right fashion, taking into consideration no dependencies other than word adjacency. 

We amend the above setting by introducing a novel layer on the top of the compositional one, which scores the linguistic plausibility of the composed sentence or phrase vector with regard to \textit{both} syntax and semantics. Following \newcite{collobert2008unified}, we convert the unsupervised learning problem to a supervised one by corrupting training sentences. Specifically, for each sentence $s$ we create two sets of negative examples. In the first set, $S'$, the target word within a given context is replaced by a random word; as in the original C\&W paper, this set is used to enforce semantic coherence in the word vectors. Syntactic coherence is enforced by a second set of negative examples, $S''$, in which the words of the context have been randomly shuffled. The objective function is defined in terms of the following hinge losses:

\vspace{-0.5cm}

\begin{equation}
\sum\limits_{s \in S}\sum\limits_{s'\in S'}\operatorname{max}(0, m-f(s)+f(s'))
\label{equ:obj1}
\end{equation}
\vspace{-0.2cm}
\begin{equation}
\sum\limits_{s \in S}\sum\limits_{s''\in S''}\operatorname{max}(0, m-f(s)+f(s''))
\label{equ:obj2}
\end{equation}
\vspace{0.1cm}

\noindent
where $S$ is the set of sentences, $f$ the compositional layer, and $m$ a margin we wish to retain between the scores of the positive training examples and the negative ones. During training, all parameters in the scoring layer, the compositional layers and word representations are jointly updated by error back-propagation. As output, we get both general-purpose syntax-aware word representations and weights for the corresponding compositional model.  

\section{From words to senses}
\label{sec:senses}

We now extend our model to address lexical ambiguity. We achieve that by applying a gated architecture, similar to the one used in the multi-sense model of \newcite{neelakantan2014efficient}, but advancing the main idea to the compositional setting detailed in Section \ref{sec:syntax}. 

\begin{figure}[t]
  \centering
  \includegraphics[scale=0.42]{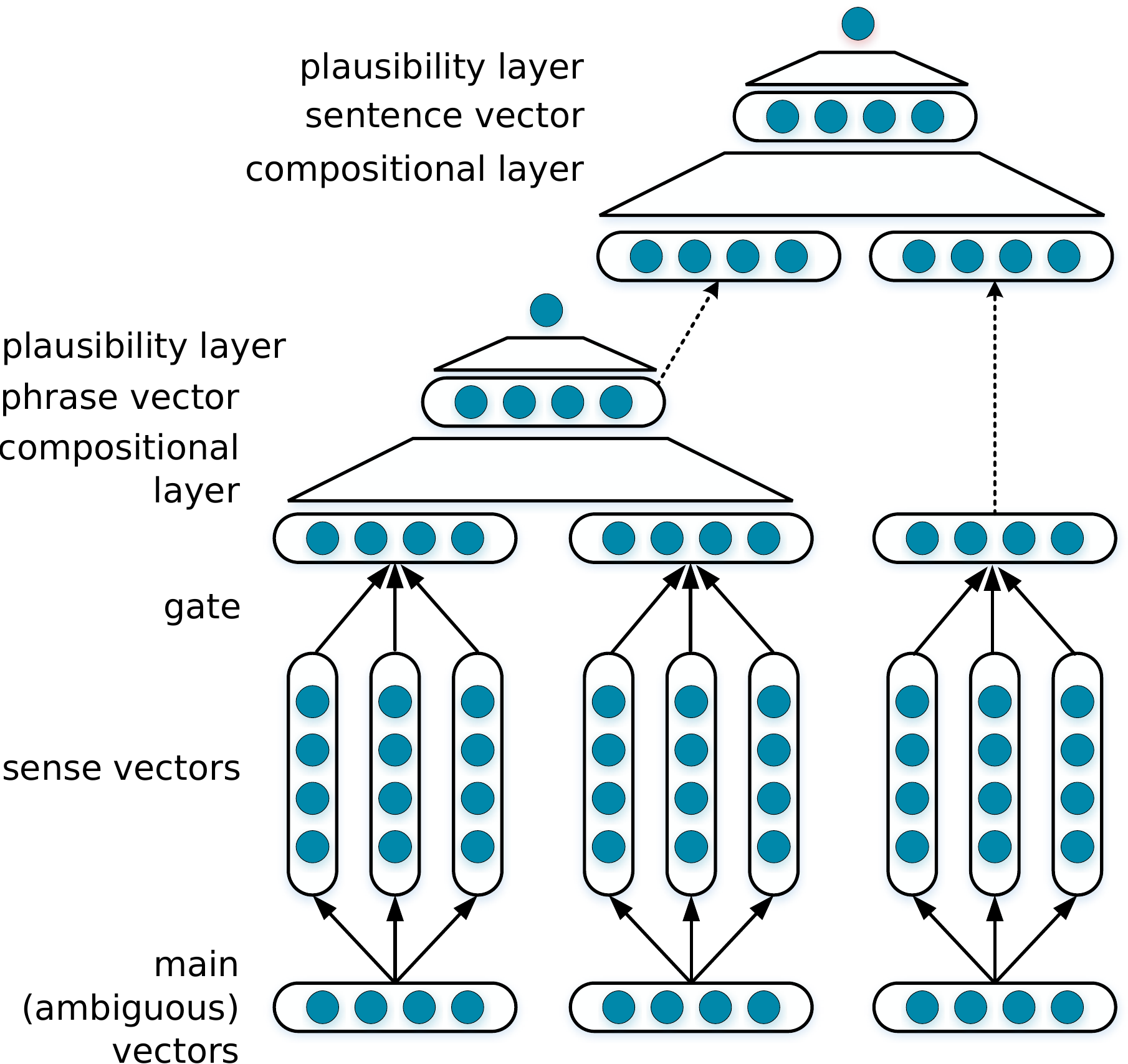}
  \caption{Training of syntax-aware multi-sense embeddings in the context of a RecNN.}
   \label{fig:overall}
\end{figure}

We assume a fixed number of $n$ senses per word.\footnote{Note that in principle the fixed number of senses assumption is not necessary; \newcite{neelakantan2014efficient}, for example, present a version of their model in which new senses are added dynamically when appropriate.} Each word is associated with a main vector (obtained for example by using an existing vector set, or by simply applying the process of Section \ref{sec:syntax} in a separate step), as well as with $n$ vectors denoting cluster centroids and an equal number of sense vectors. Both cluster centroids and sense vectors are randomly initialized in the beginning of the process. For each word $w_t$ in a training sentence, we prepare a context vector by averaging the main vectors of all other words in the same context. This context vector is compared with the cluster centroids of $w_t$ by cosine similarity, and the sense corresponding to the closest cluster is selected as the most representative of $w_t$ in the current context. The selected cluster centroid is updated by the addition of the context vector, and the associated sense vector is passed as input to the compositional layer. The selected sense vectors for each word in the sentence are updated by back-propagation, based on the objectives of Equations \ref{equ:obj1} and \ref{equ:obj2}. The overall architecture of our model, as described in this and the previous section, is illustrated in Figure \ref{fig:overall}.

\section{Task-specific dynamic disambiguation}
\label{sec:dynamic}

The model of Figure \ref{fig:overall} decouples the training of word vectors and compositional parameters from a specific task, and as a consequence from any task-specific training dataset. However, note that by replacing the plausibility layer with a classifier trained for some  task at hand, you get a task-specific network that transparently trains multi-sense word embeddings and applies dynamic disambiguation on the fly. While this idea of a single-step direct training seems appealing, one consideration is that the task-specific dataset used for the training will not probably reflect the linguistic variety that is required to exploit the expressiveness of the setting in its full. Additionally, in many cases the size of datasets tied to specific tasks is prohibiting for training a deep architecture.

\begin{figure}[b!]
\centering
\includegraphics[width=.47\textwidth]{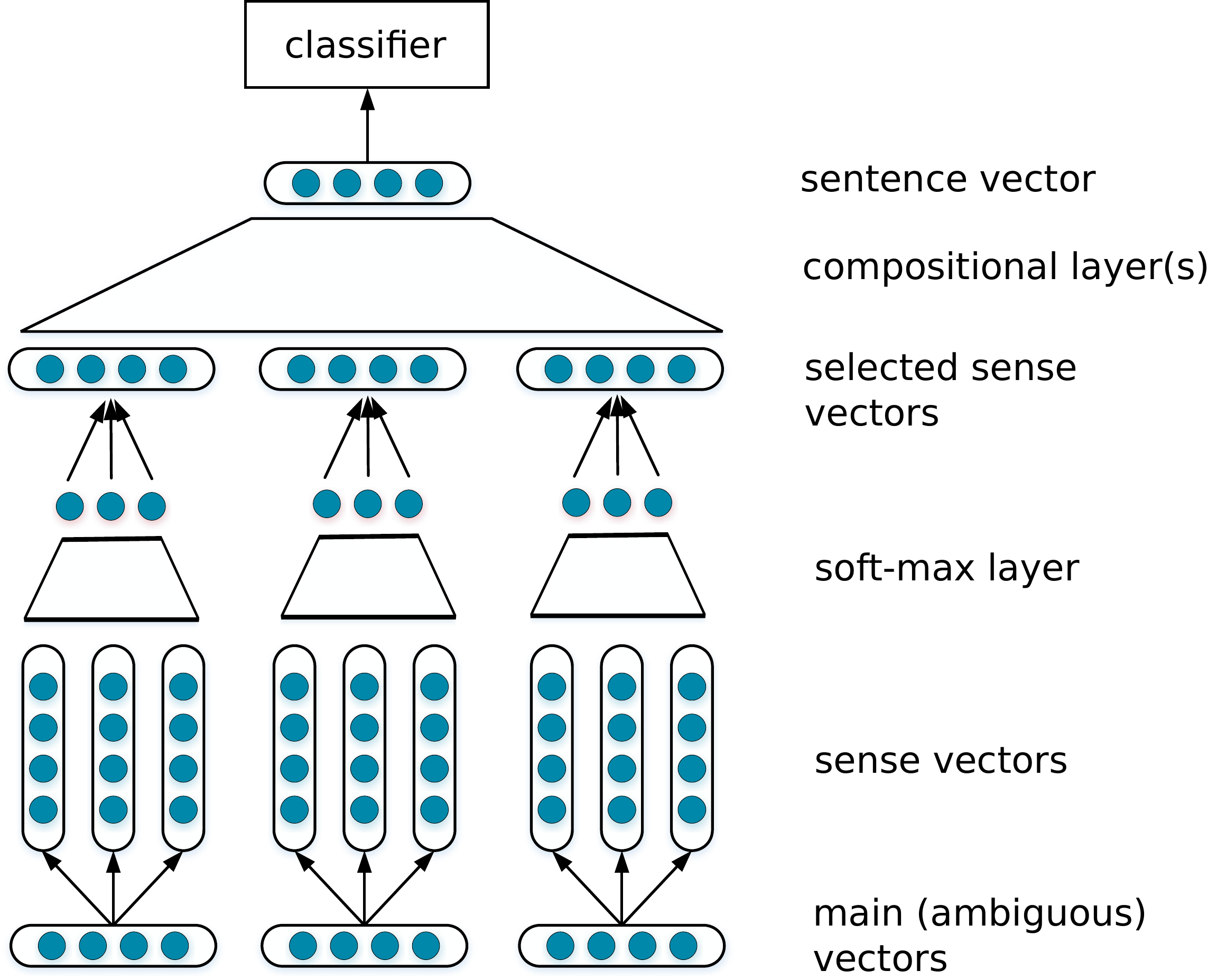}
\caption{Dynamic disambiguation in a generic compositional deep net.}
\label{fig:dd}
\end{figure}

It is a merit of this proposal that, in cases like these, it is possible for one to train the generic model of Figure \ref{fig:overall} on any large corpus of text, and then use the produced word vectors and compositional weights to initialize the parameters of a more specific version of the architecture. As a result, the trained parameters will be further refined according to the task-specific objective. Figure \ref{fig:dd} illustrates the generic case of a compositional framework applying dynamic disambiguation. Note that here sense selection takes place by a soft-max layer, which can be directly optimized on the task objective.

\section{A siamese network for paraphrase detection}
\label{sec:siamese}

We will test the dynamic disambiguation framework of Section \ref{sec:dynamic} in a paraphrase detection task. A \textit{paraphrase} is a restatement of the meaning of a sentence using different words and/or syntax. The goal of a paraphrase detection model, thus, is to examine two sentences and decide if they express the same meaning. 

While  the usual way to approach this problem is to utilize a classifier that acts (for example) on the concatenation of the two sentence vectors, in this work we follow a novel perspective: specifically, we apply a \textit{siamese architecture} \cite{bromley1993signature}, a concept that has been extensively used in computer vision \cite{hadsell2006dimensionality,sun2014deep}. While siamese networks have been also used in the past for NLP purposes (for example, by \newcite{yih-2011}), to the best of our knowledge this is the first time that such a setting is applied for paraphrase detection.

\begin{figure}[t!]
\centering
\includegraphics[scale=0.38]{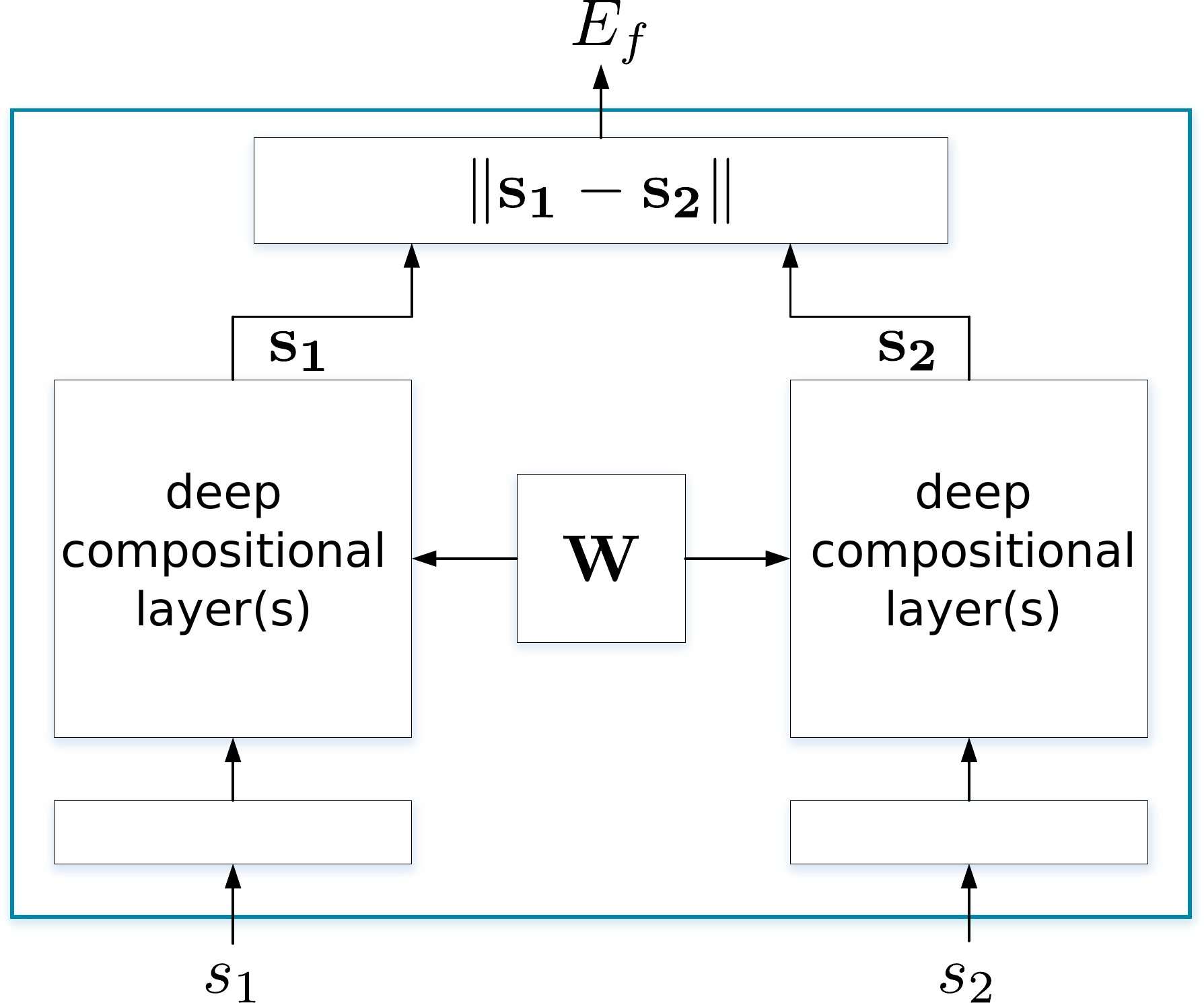}
\caption{A siamese network for paraphrase detection.}
\label{fig:siamese}
\end{figure}

In our model, two networks sharing the same parameters are used to compute the vectorial representations of two sentences, the paraphrase relation of which we wish to detect; this is achieved by employing a cost function that compares the two vectors. There are two commonly used cost functions: the first is based on the $L_2$ norm \cite{hadsell2006dimensionality,sun2014deep}, while the second on the cosine similarity \cite{nair2010rectified,sun2014deep}. The $L_2$ norm variation is capable of handling differences in the magnitude of the vectors. Formally, the cost function is defined as:

\vspace{-0.3cm}
\small
\begin{equation*}
    E_f=
    \begin{cases}
      \frac{1}{2}\left \|  f(s_1) - f(s_2) \right \|_2^2, & \text{if}\ y=1 \\
      \frac{1}{2}\operatorname{max}(0,m-\left \|  f(s_1) - f(s_2) \right \|_2)^2, & o.w.
    \end{cases}
\end{equation*}
\normalsize

\noindent  
where $s_1,s_2$ are the input sentences, $f$ the compositional layer (so $f(s_1)$ and $f(s_2)$ refer to sentence vectors), and $y=1$ denotes a paraphrase relationship between the sentences; $m$ stands for the margin, a hyper-parameter chosen in advance. On the other hand, the cost function based on cosine similarity handles only directional differences, as follows:

\begin{equation}
   E_f=\frac{1}{2}(y-\sigma(wd+b))^2
\end{equation} 

\noindent
where $d=\frac{f(s_1) \cdot f(s_2)}{\left \|  f(s_1) \right \|_2 \left \|  f(s_2) \right \|_2}$ is the cosine similarity of the two sentence vectors, $w$ and $b$ are the scaling and shifting parameters to be optimized, $\sigma$ is the sigmoid function and $y$ is the label. In the experiments that will follow in Section \ref{sec:paraphrase}, both of these cost functions are evaluated. The overall architecture is shown in Figure \ref{fig:siamese}. 

In Section \ref{sec:paraphrase} we will use the pre-trained vectors and compositional weights for deriving sentence representations that will be subsequently fed to the siamese network. When the dynamic disambiguation framework is used, the sense vectors of the words are updated during training so that the sense selection process is gradually refined.

\section{Experiments}
\label{sec:experiments}

We evaluate the quality of the compositional word vectors and the proposed deep compositional framework in the tasks of word similarity and paraphrase detection, respectively. 

\subsection{Model pre-training}
\label{sec:exp-pretraing}

In all experiments the word representations and compositional models are pre-trained on the British National Corpus (BNC), a general-purpose text corpus that contains 6 million sentences of written and spoken English. For comparison we train two sets of word vectors and compositional models, one ambiguous and one multi-sense (fixing 3 senses per word). The dimension of the embeddings is set to 300. 

As our compositional architectures we use a RecNN and an RNN. In the RecNN case, the words are composed by following the result of an external parser, while for the RNN the composition takes place in sequence from left to right. To avoid the exploding or vanishing gradient problem \cite{bengio1994learning} for long sentences, we employ a {\em long short-term memory} (LSTM) network \cite{hochreiter1997long}. During the training of each model, we minimize the hinge loss in Equations \ref{equ:obj1} and \ref{equ:obj2}. 
The plausibility layer is implemented as a 2-layer network, with 150 units at the hidden layer, and is applied at each individual node (as opposed to a single application at the sentence level). All parameters are updated with mini-batches by AdaDelta \cite{zeiler2012adadelta} gradient descent method ($\lambda=0.03$, initial $\alpha=0.05$).

\subsection{Qualitative evaluation of the word vectors}
\label{sec:qualitative}

As a first step, we qualitatively evaluate the trained word embeddings by examining the nearest neighbours lists of a few selected words. We compare the results with those produced by the skip-gram model (SG) of \newcite{mikolov2013distributed} and the language model (CW) of \newcite{collobert2008unified}. We refer to our model as SAMS (Syntax-Aware Multi-Sense). The results in Table \ref{neighbour} show clearly that our model tends to group words that are both semantically and syntactically related; for example, and in contrast with the compared models which group words only at the semantic level, our model is able to retain tenses, numbers (singulars and plurals), and gerunds. 


\begin{table*}[ht]
\footnotesize
\begin{center}
\begin{tabular}{l || L{4.2cm} | L{4.2cm} | L{4.2cm}}
\hline 
 & \bf SG & \bf CW & \bf SAMS \\ 
 \hline
 begged & beg, begging, cried & begging, pretended, beg & persuaded, asked, cried\\ 
 refused & refusing, refuses, refusal & refusing , declined, refuse & declined, rejected, denied\\ 
 interrupted & interrupting, punctuated, interrupt & interrupts, interrupt, interrupting & punctuated, preceded, disrupted\\ 
 themes & thematic, theme, notions & theme, concepts, subtext & meanings, concepts, ideas \\ 
 patiently & impatiently, waited, waits & impatiently, queue, expectantly & impatiently, siliently, anxiously \\ 
 player & players, football, league & game, club, team & athlete, sportsman, team \\ 
prompting & prompted, prompt, sparking & prompt, amid, triggered & sparking, triggering, forcing \\ 
 reproduce & reproducing, replicate, humans & reproducing, thrive, survive & replicate, produce, repopulate \\ 
predictions & prediction, predict, forecasts & predicting, assumption, predicted & expectations, projections, forecasts \\ 
\hline
\end{tabular}
\end{center}
\caption{\label{neighbour} Nearest neighbours for a number of words with various embedding models.}
\end{table*}
\normalsize

The observed behaviour is comparable to that of embedding models with objective functions conditioned on grammatical relations between words; \newcite{levy-2014}, for example, present a similar table for their dependency-based extension of the skip-gram model. The advantage of our approach against such models is twofold: firstly, the word embeddings are accompanied by a generic compositional model that can be used for creating sentence representations independently of any specific task; and secondly, the training is quite forgiving to data sparsity problems that in general a dependency-based approach would intensify (since context words are paired with the grammatical relations they occur with the target word). As a result, a small corpus such as the BNC is sufficient for producing high quality syntax-aware word embeddings.

\subsection{Word similarity}
\label{sec:similarity}

We now proceed to a quantitative evaluation of our embeddings on the Stanford Contextual Word Similarity (SCWS) dataset of \newcite{huang2012improving}. The dataset contains 2,003 pairs of words and the contexts they occur in. We can therefore make use of the contextual information in order to select the most appropriate sense for each ambiguous word. Similarly to  \newcite{neelakantan2014efficient}, we use three different metrics: \textit{globalSim} measures the similarity between two ambiguous word vectors; \textit{localSim} selects a single sense for each word based on the context and computes the similarity between the two sense vectors; \textit{avgSim} represents each word as a weighted average of all senses in the given context and computes the similarity between the two weighted sense vectors. 

We compute and report the Spearman's correlation between the embedding similarities and human judgments (Table \ref{similarity}). In addition to the skip-gram and Collobert and Weston models, we also compare against the CBOW model \cite{mikolov2013distributed} and the multi-sense skip-gram (MSSG) model of \newcite{neelakantan2014efficient}. 

\begin{table}[h]
\begin{center}
\begin{tabular}{lccc}
\hline \bf Model & \bf globalSim & \bf localSim & \bf avgSim \\ \hline
CBOW & 59.5 & -- & -- \\
SG & \textbf{61.8} & -- & -- \\
CW & 55.3  & -- & -- \\
MSSG &  61.3 & 56.7 & 62.1 \\
SAMS  & 59.9 & \textbf{58.5}  & \textbf{62.5} \\
\hline
\end{tabular}
\end{center}
\caption{\label{similarity} Results for the word similarity task (Spearman's $\rho$ $\times$ 100).}
\end{table}

Among all methods, only the MSSG model and ours are capable of learning multi-prototype word representations. Our embeddings show top performance for \textit{localSim} and \textit{avgSim} measures, and performance competitive to that of MSSG and SG for \textit{globalSim}, both of which use a hierarchical soft-max as their objective function. Compared to the original C\&W model, our version presents an improvement of 4.6\%---a clear indication for the effectiveness of the proposed learning method and the enhanced objective.

\subsection{Paraphrase detection}
\label{sec:paraphrase}

In the last set of experiments, the proposed compositional distributional framework is evaluated on the Microsoft Research Paraphrase Corpus (MSRPC) \cite{dolan2005automatically}, which contains 5,800 pairs of sentences. This is a binary classification task, with labels provided by human annotators. We apply the siamese network detailed in Section \ref{sec:siamese}.

While MSRPC is one of the most used datasets for evaluating paraphrase detection models, its size is prohibitory for any attempt of training a deep architecture. Therefore, for our training we rely on a much larger external dataset, the Paraphrase Database (PPDB) \cite{ganitkevitch2013ppdb}. The PPDB contains more than 220 million paraphrase pairs, of which 73 million are phrasal paraphrases and 140 million are paraphrase patterns that capture syntactic transformations of sentences. We use these phrase- and sentence-level paraphrase pairs as additional training contexts to fine-tune the generic compositional model parameters and word embeddings and to train the baseline models. The original training set of the MSRPC is used as validation set for deciding hyperparameters, such as the margin of the error function and the number of training epochs. 

The evaluations were conducted on various aspects, and the models are gradually refined to demonstrate performance within the state-of-the-art range. 

\paragraph{Comparison of the two error functions}
In the first evaluation, we compare the two error functions of the siamese network using only ambiguous vectors. As we can see in Table \ref{error}, the cosine error function consistently outperforms the $L_2$ norm-based one for both compositional models, providing a yet another confirmation of the already well-established fact that similarity in semantic vector spaces is better reflected by length-invariant measures.

\begin{table}[h]
\begin{center}
\begin{tabular}{lcc}
\hline \bf Model & \bf $L_2$ & \bf Cosine 
\\ \hline
RecNN & 73.8 & \textbf{74.9} \\
RNN   & 73.0 & \textbf{74.3} \\
\hline
\end{tabular}
\end{center}
\caption{\label{error} Results with different error functions for the paraphrase detection task (accuracy $\times$ 100). }
\end{table}

\paragraph{Effectiveness of disambiguation}
We now proceed to compare the effectiveness of the two compositional models when using ambiguous vectors and multi-sense vectors, respectively. Our error function is set to cosine similarity, following the results of the previous evaluation. When dynamic disambiguation is applied, we test two methods of selecting sense vectors: in the hard case the vector of the most plausible sense is selected, while in the soft case a new vector is prepared as the weighted average of all sense vectors according to probabilities returned by the soft-max layer (see Figure \ref{fig:dd}). As a baseline we use a simple compositional model based on vector addition. 

The dynamic disambiguation models and the additive baseline are compared with variations that use a simple prior disambiguation step applied on the word vectors. This is achieved by first selecting for each word the sense vector that is the closest to the average of all other word vectors in the same sentence, and then composing the selected sense vectors without further considerations regarding ambiguity. The baseline model and the prior disambiguation variants are trained as separate logistic regression classifiers. The results are shown in Table \ref{disambiguation}.


\begin{table}[h]
\small
\begin{center}
\begin{tabular}{lcccc}
\hline 
\bf Model & \bf Ambig. & \bf Prior & \bf Hard DD & \bf Soft DD \\ 

\hline
 Addition & 69.9 & 71.3 & -- & -- \\
 RecNN    & 74.9 & 75.3 & 75.7 & \bf 76.0 \\
 RNN      & 74.3 & 74.6 & 75.1 & \bf 75.2 \\
\hline
\end{tabular}
\end{center}
\normalsize
\caption{\label{disambiguation} Different disambiguation choices for the paraphrase detection task (accuracy $\times$ 100).}
\end{table}

Overall, disambiguated vectors work better than the ambiguous ones, with the improvement to be more significant for the additive model; there, a simple prior disambiguation step produces 1.4\% gains. For the deep compositional models, simple prior disambiguation is still helpful with small improvements, a result which is consistent with the findings of \newcite{cheng2014}. The small gains of the prior disambiguation models over the ambiguous models clearly show that deep architectures are quite capable of performing this elementary form of sense selection intrinsically, as part of the learning process itself. However, the situation changes when the dynamic disambiguation framework is used, where the gains over the ambiguous version become more significant. Comparing the two ways of dynamic disambiguation (hard method and soft method), the numbers that the soft method gives are slightly higher, producing a total gain of 1.1\% over the ambiguous version for the RecNN case.\footnote{For all subsequent experiments, the reported results are based on the soft selection method.}

Note that, at this stage, the advantage of using the dynamic disambiguation framework over simple prior disambiguation is still small (0.7\% for the case of RecNN). We seek the reason behind this in the recursive nature of our architecture, which tends to progressively ``hide'' local features of word vectors, thus diminishing the effect of the fine-tuned sense vectors produced by the dynamic disambiguation mechanism. The next section discusses the problem and provides a solution.

\paragraph{The role of pooling}
One of the problems of the recursive and recurrent compositional architectures, especially in grammars with strict branching structure such as in English, is that any given composition is usually the product of a terminal and a non-terminal; i.e. a single word can contribute to the meaning of a sentence to the same extent as the rest of a sentence on its whole, as below:

\vspace{0.2cm}
$[[$kids$]_{\text{NP}}$ $[$play ball games in the park$]_{\text{VP}}]_\text{S}$
\vspace{0.2cm}

In the above case, the contribution of the words within the verb phrase to the final sentence representation will be faded out due to the recursive composition mechanism. Inspired by related work in computer vision \cite{sun2014deep}, we attempt to alleviate this problem by introducing an average pooling layer at the sense vector level and adding the resulting vector to the sentence representation. By doing this we expect that the new sentence vector will reflect local features from all words in the sentence that can help in the classification in a more direct way. The results for the new deep architectures are shown in Table \ref{pooling}, where we see substantial improvements for both deep nets. More importantly, the effect of dynamic disambiguation now becomes more significant, as expected by our analysis. 

Table \ref{pooling} also includes results for two models trained in a single step, with word and sense vectors randomly initialized at the beginning of the process. We see that, despite the large size of the training set, the results are much lower than the ones obtained when using the pre-training step. This demonstrates the importance of the initial training on a general-purpose corpus: the resulting vectors reflect linguistic information that, although not obtainable from the task-specific training, can make great difference in the result of the classification.

\begin{table}[h]
\small
\begin{center}
\begin{tabular}{lccc}
\hline 
\bf Model & \bf Ambig. & \bf Prior & \bf Dynamic \\ 
\hline
 RecNN+pooling & 75.5 & 76.3 & \bf 77.6 \\
 RNN+pooling   & 74.8 & 75.9 & \bf 76.6  \\
\hline
 1-step RecNN+pooling & 74.4 & -- & 72.9 \\ 
 1-step RNN+pooling   & 73.6 & -- & 73.1 \\
\hline
\end{tabular}
\end{center}
\normalsize
\caption{\label{pooling} Results with average pooling for the paraphrase detection task (accuracy $\times$ 100). }
\end{table}

\paragraph{Cross-model comparison}
In this section we propose a method to further improve the performance of our models, and we present an evaluation against some of the previously reported results. 

We notice that using distributional properties alone cannot capture efficiently subtle aspects of a sentence, for example numbers or human names. However, even small differences on those aspects between two sentences can lead to a different classification result. Therefore, we train (using the MSPRC training data) an additional logistic regression classifier which is based not only on the embeddings similarity, but also on a few hand-engineered features. We then ensemble the new classifier (C1) with the original one. In terms of feature selection, we follow \newcite{socher2011dynamic} and \newcite{blacoe2012comparison} and add the following features: the difference in sentence length, the unigram overlap among the two sentences, features related to numbers (including the presence or absence of numbers from a sentence and whether or not the numbers in the two sentences are the same). In Table \ref{model comparison} we report results of the original model and the ensembled model, and we compare with the performance of other existing models.

\begin{table}[t!]
\small
\begin{center}
\begin{tabular}{l|lcc}
\hline 
& \bf Model  & \bf Acc. & \bf F1\\ 
\hline\hline
\multirow{2}{*}{\begin{sideways}\bf BL\end{sideways}} & 
All positive & 66.5 & 79.9  \\
& Addition (disamb.) & 71.3  & 81.1 \\ \hline
\multirow{6}{*}{\begin{sideways}\textbf{Dynamic Dis.}\end{sideways}} &
RecNN & 76.0 & 84.0  \\
& RecNN+Pooling &  77.6 &  84.7 \\
& RecNN+Pooling+C1 &  \bf 78.6 &\bf  85.3 \\
& RNN  &  75.2 & 83.6  \\
& RNN+Pooling & 76.6 &  84.3 \\
& RNN+Pooling+C1 & 77.5 & 84.6  \\ \hline
\multirow{9}{*}{\begin{sideways}\bf{Published results}\end{sideways}} & 
\newcite{mihalcea2006corpus} & 70.3 & 81.3 \\
& \newcite{rus2008paraphrase} & 70.6 & 80.5 \\
& \newcite{qiu2006paraphrase} & 72.0 & 81.6 \\
& \newcite{islam2009semantic}  & 72.6 & 81.3 \\
& \newcite{fernando2008semantic} & 74.1 & 82.4 \\
& \newcite{wan2006using} & 75.6 & 83.0 \\
& \newcite{das2009paraphrase} & 76.1 & 82.7 \\
& \newcite{socher2011dynamic} & 76.8 & 83.6 \\
& \newcite{madnani2012re} & 77.4 & 84.1 \\ 
& \newcite{ji-2013} & \bf{80.4} & \bf{85.9} \\
\hline
\end{tabular}
\end{center}
\normalsize
\caption{\label{model comparison} Cross-model comparison in the paraphrase detection task.}
\end{table}

In all of the implemented models (including the additive baseline), disambiguation is performed to guarantee the best  performance. We see that by ensembling the original classifier with C1, we improve the result of the previous section by another 1\%. This is the second best result reported so far for the specific task, with a 0.6 difference in F-score from the first \cite{ji-2013}.\footnote{Source: ACL Wiki ({\tt\fontsize{8}{8}\selectfont http://www.aclweb.org/acl- wiki}), August 2015.}

\section{Conclusion and future work}

The main contribution of this paper is a deep compositional distributional model acting on linguistically motivated word embeddings.\footnote{Code in Python/Theano and the word embeddings can be found at  {\tt\fontsize{8}{8}\selectfont https://github.com/cheng6076}.} The effectiveness of the syntax-aware, multi-sense word vectors and the dynamic compositional disambiguation framework in which they are used was demonstrated by appropriate tasks at the lexical and sentence level, respectively, with very positive results. As an aside, we also demonstrated the benefits of a siamese architecture in the context of a paraphrase detection task. While the architectures tested in this work were limited to a RecNN and an RNN, the ideas we presented are in principle directly applicable to any kind of deep network. As a future step, we aim to test the proposed models on a convolutional compositional architecture, similar to that of \newcite{KalchbrennerACL2014}.

\section*{Acknowledgments}

The authors would like to thank the three anonymous reviewers for their useful comments, as well as Nal Kalchbrenner and Ed Grefenstette for early discussions and suggestions on the paper, and Simon \v{S}uster for comments on the final draft. Dimitri Kartsaklis gratefully acknowledges financial support by AFOSR.

\bibliographystyle{acl}
\bibliography{emnlp2015}
\end{document}